\PassOptionsToPackage{table}{xcolor}
\documentclass[]{fairmeta}
\usepackage{amssymb}
\usepackage{multirow}
\usepackage{bigdelim}
\usepackage{todonotes}
\usepackage{longtable}
\usepackage{tabularray}
\usepackage{wrapfig}
\usepackage[most]{tcolorbox} 
\usepackage{url}
\usepackage{float}
\usepackage{multirow}      
\usepackage{arydshln}      
\usepackage{graphicx}      
\usepackage{makecell} 
\usepackage{colortbl}   
\usepackage{booktabs}   

\definecolor{headergray}{gray}{0.92}
\definecolor{rowblue0}{rgb}{0.97, 0.97, 1.0}
\definecolor{rowblue1}{HTML}{EBF4FF} 
\definecolor{rowblue2}{HTML}{E0E9F5}
\definecolor{rowblue3}{HTML}{D5E0F0}

\newcommand{\model}{\texttt{ARC-Hunyuan-Video}}

\newcommand{\modelbase}{\model\texttt{-7B}}

\definecolor{prompt}{HTML}{5f84e4}
\definecolor{img}{HTML}{820100}

\title{ARC-Hunyuan-Video-7B: Structured Video Comprehension of Real-World Shorts}

\author[1,*]{Yuying Ge}
\author[1,*,\dagger]{Yixiao Ge}
\author[1,*]{Chen Li}
\author[1,*]{Teng Wang}
\author[1,*]{Junfu Pu}
\author[1,*]{Yizhuo Li}
\author[1,*]{Lu Qiu}
\author[2]{Jin Ma}
\author[2]{Lisheng Duan}
\author[2]{Xinyu Zuo}
\author[2]{Jinwen Luo}
\author[3]{Weibo Gu}
\author[4]{Zexuan Li}
\author[2]{Xiaojing Zhang}
\author[3]{Yangyu Tao}
\author[3]{Han Hu}
\author[3]{Di Wang}
\author[1]{Ying Shan}

\affiliation[1]{ARC Lab, Tencent PCG}
\affiliation[2]{Search Application Department, Tencent CSIG}
\affiliation[3]{Tencent Hunyuan}
\affiliation[4]{Big Data Platform Department, Tencent PCG}

\contribution[*]{\tt Core contributors}
\contribution[\dagger]{\tt Project lead}

\abstract{
Real-world user-generated short videos, especially those distributed on platforms such as WeChat Channel and TikTok, dominate the mobile internet. However, current large multimodal models lack essential temporally-structured, detailed, and in-depth video comprehension capabilities, which are the cornerstone of effective video search and recommendation, as well as emerging video applications.
Understanding real-world shorts is actually challenging due to their complex visual elements, high information density in both visuals and audio, and fast pacing that focuses on emotional expression and viewpoint delivery. This requires advanced reasoning to effectively integrate multimodal information, including visual, audio, and text. 
In this work, we introduce \model\footnote{The version supports Chinese and English videos and particularly excels at Chinese.}, a multimodal model that processes visual, audio, and textual signals from raw video inputs end-to-end for structured comprehension. 
The model is capable of multi-granularity timestamped video captioning and summarization, open-ended video question answering, temporal video grounding, and video reasoning.
Leveraging high-quality data from an automated annotation pipeline, our compact 7B-parameter model is trained through a comprehensive regimen: pre-training, instruction fine-tuning, cold start, reinforcement learning (RL) post-training, and final instruction fine-tuning. 
Quantitative evaluations on our introduced benchmark ShortVid-Bench and qualitative comparisons demonstrate its strong performance in real-world video comprehension, and it supports zero-shot or fine-tuning with a few samples for diverse downstream applications. The real-world production deployment of our model has yielded tangible and measurable improvements in user engagement and satisfaction, a success supported by its remarkable efficiency, with stress tests indicating an inference time of just 10 seconds for a one-minute video on H20 GPU\footnote{The reported inference time is based on a deployment accelerated with vLLM.}.

}

\date{July 28, 2025}

\metadata[Code]{\url{https://github.com/TencentARC/ARC-Hunyuan-Video-7B}}

\begin{document}
\maketitle

\begin{figure}[h!]
\centering
\makebox[\textwidth][c]{\includegraphics[width=1.0\linewidth]{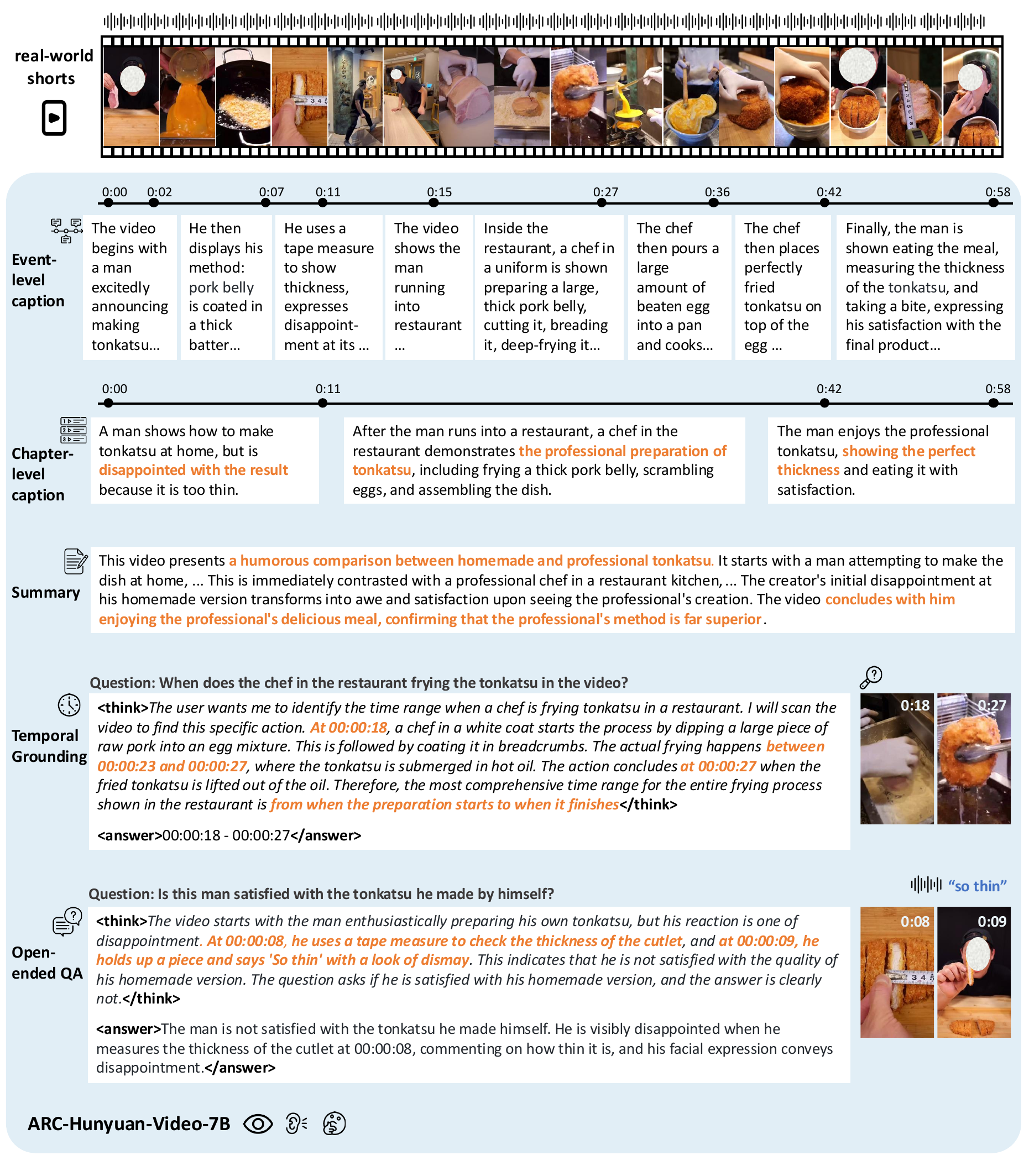}}%
\caption{Model capabilities of \modelbase, which supports multi-granular timestamped captioning (output time span and corresponding description), summarization, temporal grounding, and open-ended question answering through integrating and reasoning over both visual and audio cues in the user-generated short videos.}
\label{fig:teaser_example}
\vspace{10pt}
\end{figure}

\section{Introduction}
The explosive growth of user-generated short videos on platforms such as WeChat Channel and TikTok has fundamentally reshaped mobile internet consumption. These short videos, characterized by their brevity, diversity, and high engagement, have become a dominant medium for information sharing, entertainment, and social interaction. This shift has created an urgent need for structured, temporally-aware, and in-depth comprehension of real-world video content, which is essential for enabling a wide range of video-centric applications, including search, recommendation, and emerging intelligent video services.

Yet, the very characteristics that make these videos so appealing also present significant challenges for automated understanding. These videos typically contain dense visual elements (e.g., visual effects, text overlays), high-information audios (including speech and sound effects), and rapid pacing that emphasizes emotional expression and viewpoint delivery. Effectively comprehending such complex content requires not only multimodal integration of visual, audio, and textual information, but also advanced reasoning capabilities to grasp the core intent of the content. However, current multimodal models, which are primarily designed for general video understanding, struggle to address these challenges and fall short of delivering the level of comprehension required for real-world applications.

To bridge this gap, we introduce \model, a multimodal model for comprehensive understanding of real-world short videos. Our model processes visual, audio, and textual inputs to achieve what we term \textbf{Structured Video Comprehension}: the ability to decompose a video into its constituent events and narrative elements with temporal precision. This includes generating multi-granularity timestamped video caption and summary, answering open-ended questions through video reasoning, and performing accurate temporal grounding of events as illustrated in Fig.~\ref{fig:teaser_example}. This structured understanding is crucial for real-world scenarios, as it allows the model to move beyond surface-level analysis and truly understand what happens in user-generated content, when it happens, why it matters, and what intentions the creator wanted to convey.

Specifically, \model\ is built on top of the Hunyuan-7B vision-language model and has two key incremental designs to meet the requirements of effective structured video comprehension:
(1) an extra audio encoder with fine-grained visual-audio synchronization for temporally aligned visual-audio inputs, 
and (2) a timestamp overlay mechanism on visual frames that explicitly provides the model with temporal awareness for accurate event localization. 
Additionally, we collect millions of in-house real-world videos and develop a totally automated bootstrapped annotation pipeline that generates high-quality data, enabling a comprehensive training regimen. 
This includes (i) pre-training for fundamental knowledge and atomic capability acquisition, (ii) instruction fine-tuning for task alignment, (iii) cold-start initialization, (iv) reinforcement learning (RL) post-training, and (v) final instruction fine-tuning using high-quality human-annotated data and trajectories obtained through rejection sampling.
A key aspect of our RL strategy is the design of objective questions such as multiple-choice questions and temporal grounding to enhance the model's holistic comprehension of the video. This design is grounded in our pilot experiments, which demonstrate that verifiable tasks with RL significantly benefits the learning of high-quality subjective data (e.g., video summary).

To rigorously evaluate our model’s ability to understand real-world short videos, we construct a specialized, human-annotated benchmark named \textbf{ShortVid-Bench} with multiple-choice questions. Empirical evaluations demonstrate that our compact 7B-parameter model not only achieves exceptional performance in real-world video understanding on our proposed benchmark, but also excels in temporal video grounding benchmarks. Furthermore, \model\ exhibits strong versatility, supporting zero-shot inference for a range of tasks and adapting to various downstream applications, such as video abstract for search and tagging for recommendation, with minimal fine-tuning data required. The deployment of our fine-tuned model in real-world product scenarios has resulted in significant and measurable improvements in user engagement and satisfaction. This success is underpinned by the model's remarkable efficiency: stress test reports show an inference time of just 10 seconds for a one-minute video on NVIDIA H20 GPU, yielding an average of 500 tokens, with inference accelerated by the vLLM~\citep{kwon2023efficient} framework.

To facilitate further research and application, we have open-sourced both the model checkpoint, API, and inference code. We hope that \model\ will contribute to advancing the field of structured video comprehension and inspire new developments in the comprehension of real-world short videos.

\section{Related Work}
Short-form video has emerged as a dominant medium for communication, entertainment, and information dissemination across social media platforms. The ability to automatically understand such content is crucial for a wide range of downstream applications, including content retrieval, personalized recommendation, automated video tagging, and content moderation.
Despite its importance, comprehending real-world short videos presents unique challenges. These videos are characterized by dense visual elements (e.g., dynamic effects, text overlays), rich audio streams (speech, music, sound effects), and rapid narrative pacing that emphasizes emotional expression and viewpoint delivery. These characteristics necessitate joint modeling of visual, audio, and textual modalities with advanced reasoning to comprehend key events, temporal relationships, and decipher underlying intentions.

Recent efforts have sought to address these challenges. Our concurrent work Keye-VL-8B~\citep{team2025kwai} introduces a multimodal foundation model designed specifically for short-video understanding. However, Keye-VL-8B does not directly integrate raw audio signals; instead, it relies on transcripts generated by automatic speech recognition (ASR). This approach discards important non-speech audio cues, such as emotional tone and environmental sounds, and can lead to temporal misalignment between audio and visual content.
Meanwhile,  audio-visual LLMs~\citep{shu2023audio,zhang2023video,sun2024video,sun2025video,xu2025qwen2,tang2025video} have been developed to jointly process video, audio, and text input, but they focus on video understanding of general scenarios, which feature slower pacing, and lower information density. As a result, they often struggle to capture the dynamic, fast-paced, and information-rich nature of user-generated short videos.

In this work, we propose \model, a compact 7B-parameter multimodal model that achieves structured video comprehension of real-world shorts through synchronized audio-visual-text processing and sophisticated reasoning.

\section{Method}

\begin{figure}[h!]
\centering
\makebox[\textwidth][c]{\includegraphics[width=1.0\linewidth]{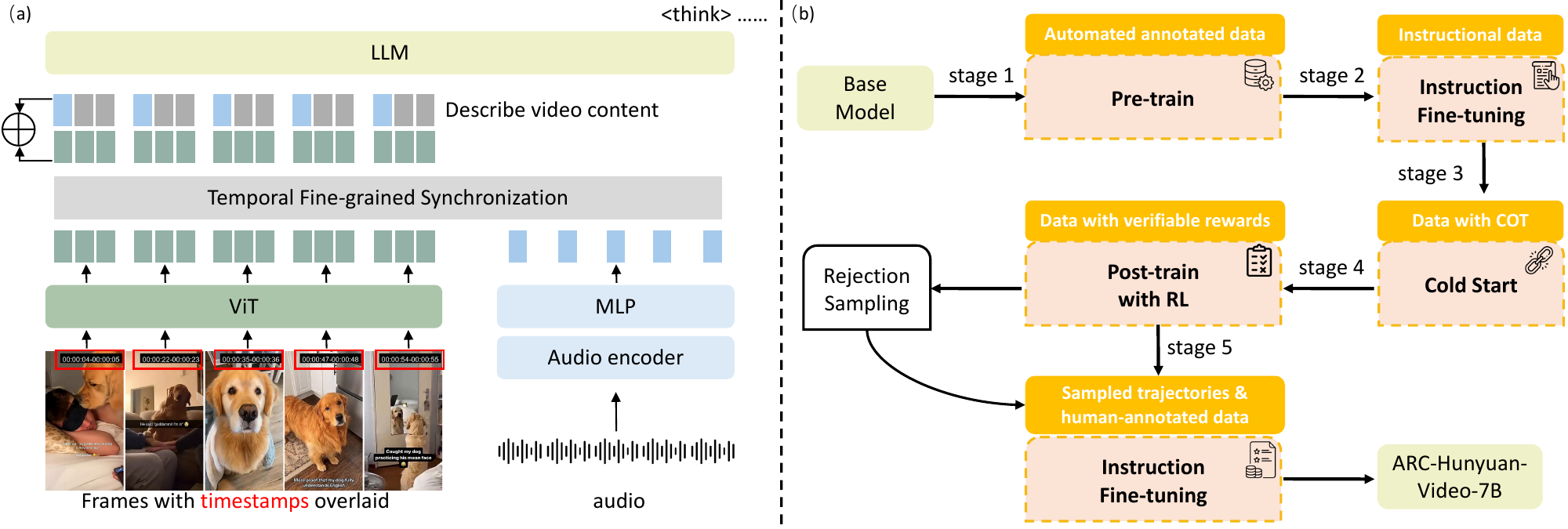}}%
\caption{(a) Model architecture. Built upon the Hunyuan-7B VLM, we incorporate an audio encoder with fine-grained visual-audio synchronization to obtain temporally aligned multimodal inputs. Timestamps are overlaid on visual frames to provide the model with temporal awareness.(b) Training stages including pre-training, instruction fine-tuning, cold start initialization, RL post-training and final instruction fine-tuning using high-quality human-annotated data and trajectories selected via rejection sampling.}
\label{fig:model}
\end{figure}

\subsection{Model Architecture}
As shown in Fig.~\ref{fig:model}, \model\ is built upon the Hunyuan-7B vision-language model (VLM), with an additional audio encoder and fine-grained visual-audio synchronization mechanism to obtain temporally aligned multimodal features as input to the LLM. To explicitly provide the model with temporal awareness, we directly overlay the timestamp of each sampled frame onto its corresponding visual frame.

\paragraph{\textbf{Visual Encoding.}}
To process the visual input, we first sample frames at a rate of one frame per second (1 fps). For videos exceeding 150 seconds, we uniformly sample a total of 150 frames to maintain a manageable sequence length. To enhance model's temporal awareness, we explicitly render the corresponding timestamp of each frame in an HH:MM:SS format onto its top-right corner. This provides the model with a direct, explicit signal for temporal localization.These timestamped frames are then fed into a pre-trained Vision Transformer (ViT)~\citep{dosovitskiy2020image} encoder. While the Hunyuan ViT architecture inherently supports dynamic input resolutions, we resize all frames to a fixed resolution of 640×640 for enabling the visual-audio synchronization detailed below. For each input frame, the ViT encoder outputs a sequence of 112 visual tokens.

\paragraph{\textbf{Audio Encoding.}}
For the audio modality, we leverage the pre-trained audio encoder from Whisper~\citep{radford2023robust}. The raw audio waveform is first segmented into 30-second chunks. The encoder processes each chunk to produce a sequence of 1500 feature tokens. For videos longer than 300 seconds, we split the audio into exactly 150 segments and truncate each to the initial 2 seconds before encoding—a design choice that optimizes temporal synchronization with visual frames. Finally, the audio features extracted by the audio encoder are passed through a multi-layer perceptron (MLP). This projection layer aligns the dimensionality of the audio features with that of the visual tokens, preparing them for fusion.

\paragraph{\textbf{Visual-audio Synchronization.}}
The fine-grained synchronization aims to obtain multimodal representation where visual and audio tokens that correspond to the same time interval are fused, ensuring that the LLM receives temporally aligned multimodal signals. We adopt an adaptive and parameter-free synchronization strategy based on video duration. Specifically, for each sampled video frame, we align and fuse the corresponding audio segment by zero-padding the audio tokens to match the number of visual tokens, then adding them to form synchronized multimodal embeddings. This approach ensures that, regardless of video length, each fused embedding consistently represents the same temporal interval. The resulting sequence of synchronized embeddings, with positional encodings added, is then input to the LLM.

\subsection{Pre-training}
\begin{figure}[h!]
\centering
\makebox[\textwidth][c]{\includegraphics[width=1.0\linewidth]{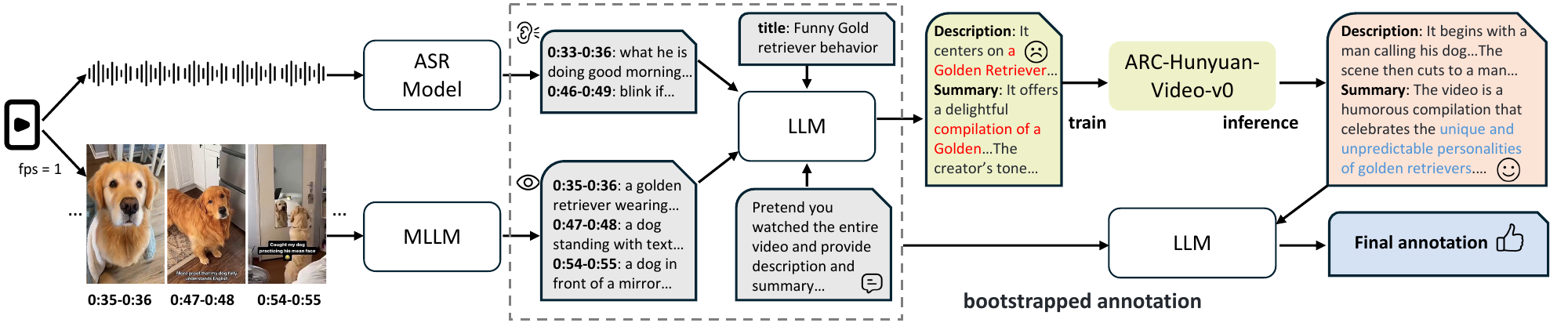}}%
\caption{Our automated bootstrapped annotation pipeline for pre-training. It extracts timestamped speech via ASR model and frame-level descriptions via MLLM; these, along with meta information (e.g., title), are input to an LLM for initial video annotation. The annotated data is used to train an initial version of the model, whose inference results are further integrated to produce the final annotations.}
\label{fig:data}
\end{figure}

\subsubsection{Annotation pipeline} To generate high-quality video descriptions and summaries that capture the essence of real-world short videos (i.e., ``truly understanding'' the content), we designed an automated bootstrapped annotation pipeline as shown in Fig.~\ref{fig:data}. This pipeline is structured to iteratively refine annotations through a self-improving process, ensuring robust and accurate multimodal integration.

We begin by extracting and integrating multimodal information from videos using a series of specialized models. Specifically, we employ Whisper-v3~\citep{radford2023robust} to transcribe speech with precise timestamps, providing synchronized audio-text data. Concurrently, we use InternVL-2.5-8B~\citep{chen2024expanding} to generate detailed captions and detect textual overlays (e.g., on-screen text) for sampled video frames. These outputs, combined with video metadata such as titles, are fed into a closed-source large language model (LLM) for comprehensive synthesis. To ensure the model ``truly understands'' the video, we implement a Chain-of-Thought (COT) prompting strategy. This guides the LLM to first output intermediate elements: a step-by-step description of events, the creator's attitude, and potential audience-interest tags. These elements are then integrated into a final summary that encapsulates the video's core intent, emotional expression, and viewpoint delivery.

Leveraging the initial annotations, we train a preliminary version of our model. This model is then deployed to generate new descriptions and summaries, which are combined with the outputs from the initial pipeline (e.g., speech transcriptions, frame captions, and metadata). The aggregated data is reprocessed through the same closed-source LLM for polishing, using the COT approach to resolve inconsistencies and enrich detail. This iterative loop, where the model's own outputs refine the annotations, yields high-quality, final descriptions and summaries that are used for pre-training. 

\subsubsection{Pre-training data} 
\paragraph{\textbf{Video description and summary.}} Using our automated bootstrapped annotation pipeline, we annotate 4.5M short-form videos with detailed descriptions and summaries. Additionally, to ensure general video understanding capabilities, we apply the same pipeline to annotate 0.2M publicly available academic videos.

\paragraph{\textbf{Image caption and OCR.}} Frame-level image understanding, including captioning and OCR, is one of the fundamental capabilities for short-form video comprehension. Therefore, we leverage the frame-level captions and OCR results obtained during the video annotation process as training data, resulting in a total of 4.7M image-text pairs.

\paragraph{\textbf{ASR.}} Automatic speech recognition (ASR) is another fundamental capability for understanding short-form videos. Therefore, we utilize the ASR results obtained during the video annotation process as training data. We further employ an LLM to filter out transcribed speech samples without meaningful semantics (retaining only a small portion and labeling them as ``no speech detected''), resulting in a total of 3.2M audio-text pairs.

\paragraph{\textbf{Video temporal grounding.}} Precise temporal video grounding, aligning textual queries with specific temporal segments within a video, is essential for structured comprehension, which improves the model’s temporal awareness. To develop this capability, we leverage a diverse collection of 0.5M temporally grounded instances sourced from multiple public datasets. These instances provide paired textual queries (comprising natural language descriptions or questions) and their corresponding temporal intervals within the videos.

\paragraph{\textbf{Video multi-granular caption.}} To support multi-granular video captioning, our dataset includes both event-level and chapter-level captions, each paired with their corresponding time spans and descriptive texts. For event-level captions, we leverage timestamped annotations from public datasets, applying filters to remove videos with excessive segment overlap or insufficient caption coverage relative to video duration, resulting in 50K high-quality samples. For chapter-level captions, we automatically generate captions for the time spans in 80K in-house videos, where the time spans themselves were annotated by humans.

\subsubsection{Training Recipe}
Building upon the pre-trained Hunyuan-7B VLM with established visual understanding capabilities, our training adopts a progressive two-stage strategy to integrate audio modality while preserving core competencies.

In the first stage, we conduct warm-up training using Automatic Speech Recognition (ASR) data to adapt the model to audio feature inputs as shown in Fig.~\ref{fig:model}. To prevent degradation of existing visual understanding, this stage simultaneously incorporates image-text pair data. When a modality is missing, we feed an all-zero input into the corresponding modality encoder. The dual-task design ensures the model develops initial audio-text alignment while retaining its foundational visual understanding abilities.

In the second stage, we perform full multimodal pre-training (video/audio/text) via next-token prediction. We freeze parameters of both the ViT and audio encoder to preserve their feature extraction capabilities. Only the MLP adapter layers and the full LLM backbone are updated. This phase employs a learning rate of 2e-5, leverages DeepSpeed Zero Stage 1 optimization, and operates with an extended context length of 20K tokens.

\subsection{Post-training}
The post-training stage aims to further enhance the structured comprehension capability of \model\ for real-world shorts. Since our automated annotation pipeline inevitably introduces noise in the generated data, we collect a small set of high-quality human-annotated data. We conduct pilot experiments to investigate how to fully leverage this high-quality data to enhance the model’s ability for truly understanding videos.

\subsubsection{Pilot Experiments}
To quantitatively evaluate our model's capability for video summarization, we curate 140 real-world shorts with human-annotated summaries and employ LLM-as-a-judge scoring (scale: 1-10) comparing model outputs against human annotations. Initial experiments revealed that supervised fine-tuning the pre-trained model directly on this human-annotated summary data yielded no significant performance gains (pretrained 6.42 vs. fine-tuned 6.67).  We further explored using Direct Preference Optimization (DPO)~\citep{rafailov2023direct}, treating the human annotations as positive samples and model outputs as negative samples, but similarly observed no improvement (pretrained 6.42 vs. DPO-tuned 6.50). We hypothesize that this stems from a potential distribution mismatch between the human annotations and the model's learned representations.

Inspired by the success of approaches like DeepSeek-R1~\citep{guo2025deepseek}, which utilized rule-based rewards on tasks with verifiable outputs (e.g., mathematics, coding) with Generalized Reinforcement Policy Optimization (GRPO) algorithm, we design two objective tasks for structured video understanding: (1) Multi-dimensional Multiple-Choice QA covering five critical aspects: (a) spatial fine-grained understanding, (b) temporal fine-grained understanding, (c) timeline analysis, (d) intent comprehension (creator's attitude/purpose), and (e) event reasoning; (2) Temporal Video Grounding. We found that conducting GRPO-based post-training on these verifiable tasks, followed by fine-tuning on the human-annotated summaries, led to substantial gains in comprehension performance (pretrained 6.42 vs. GRPO-sft 6.99). The MCQ task enhances the model’s understanding across multiple respects by explicitly targeting diverse dimensions of video comprehension, while the grounding task increases the model’s temporal awareness by requiring precise localization of events within the video timeline. This combination enables the model to develop a more holistic and temporally sensitive understanding of video content, thereby better preparing the enhanced model to effectively learn from high-quality data.

After verifying that GRPO post-training on verifiable tasks followed by fine-tuning strengthens model's comprehension ability, we design a comprehensive post-training regimen including an initial instruction fine-tuning for instruction alignment, a cold-start phase to initialize the model for reinforcement learning, a targeted reinforcement learning phase using GRPO, and a final instruction fine-tuning stage using high-quality human-annotated data.

\subsubsection{Stage 1: Initial Instruction Fine-tuning}
The Instruction Fine-tuning stage is adopted to equip \model\ with robust instruction-following capabilities. To achieve this, we construct a comprehensive and high-quality supervised dataset that covers a wide range of tasks. Specifically, our data includes 460K open-ended question answering (QA) samples and 70K multiple-choice QA from publicly available academic datasets, as well as 20K QA samples collected from real-world short videos, ensuring both general coverage and domain-specific relevance. For temporal video grounding, we incorporate 10K samples from academic datasets and 5K samples from real-world short videos, enabling precise event localization within diverse video content. Additionally, the dataset contains 45K video description and summarization samples from real-world videos, along with 12K multi-granular captioning samples. Similar to pre-training, only the MLP adapter layers and the full LLM backbone are updated during this stage. We use a learning rate of 1e-5, leverage DeepSpeed ZeRO Stage 1 optimization, and operate with an extended context length of 20K tokens.

\subsubsection{Stage 2: Cold Start Initialization for Reinforcement Learning}
To prepare a strong initial policy for the subsequent reinforcement learning phase, we fine-tune the model on a curated dataset of 146K samples featuring Chain-of-Thought (CoT) reasoning. This ``cold start'' stage aims to teach the model how to perform step-by-step reasoning across a broad spectrum of tasks, thereby building a versatile reasoning foundation. For 90K multiple-choice QA samples covering both general video understanding and real-world shorts scenarios, we use a powerful MLLM to generate CoT rationales, retaining only instances where the final answer was correct. Similarly, for 18K temporal grounding tasks again spanning general videos and real-world shorts, we generate CoT rationales for timestamp prediction and filtered for samples with an Intersection-over-Union (IoU) above 0.6 between the predicted and the ground truth time span. This process was extended to 20K open-ended QA samples, where an LLM judge verifies the correctness of the final answer, and to 15K video summarization and 3K chapter-level captioning tasks, where event-level captions serve as the intermediate reasoning steps. This model, trained with the same recipe as the initial instruction fine-tuning stage, serves as the starting point for reinforcement learning.

\subsubsection{Stage 3: Reinforcement Learning with GRPO}

Our pilot experiments demonstrate that effective reinforcement learning on verifiable tasks benefits the learning of high-quality subjective data. Therefore, the phase narrows its focus to the two tasks with objective, verifiable reward signals: 100K multiple-choice questions and 35K temporal video grounding instances. For multiple-choice questions, a binary reward of 1.0 is assigned for correct answers and 0.0 for incorrect ones. For temporal grounding tasks, the reward is determined by the Intersection over Union (IoU) between the predicted time span and the ground truth. Using the GRPO~\citep{guo2025deepseek} algorithm, we exclusively fine-tune the parameters of the large language model (LLM). Training employs a learning rate of 2e-5, leverages DeepSpeed ZeRO Stage 3 optimization, and operates with an extended context length of 20K tokens. The KL divergence coefficient within the GRPO algorithm is set to 0.1.

\subsubsection{Stage 4: Final Instruction Fine-tuning}

In the final stage, we return to the high-quality, human-annotated data. Having undergone the targeted reasoning enhancement of GRPO, the model is now capable of effectively learning from this nuanced data. 
We use a dataset of 25K human-annotated subjective questions for instruction fine-tuning including open-ended QA, video summarization and chapter-level captioning. We further leverage the enhanced capabilities of the GRPO-tuned model to generate 100K high-quality multiple-choice questions with CoT rationales and 50K temporal grounding instances with detailed reasoning traces through rejection sampling. By combining the high-quality human annotations with accurate self-generated trajectories, this final instruction fine-tuning stage polishes the model's capabilities, aligning it closely with human-level comprehension of real-world short videos. The training employs a learning rate of 1e-5 and utilizes DeepSpeed ZeRO Stage 1 optimization.

\section{Experiments}
\subsection{Qualitative Evaluation}
\subsubsection{Model Capability}
\begin{figure}[H]
\centering
\makebox[\textwidth][c]{\includegraphics[width=1.0\linewidth]{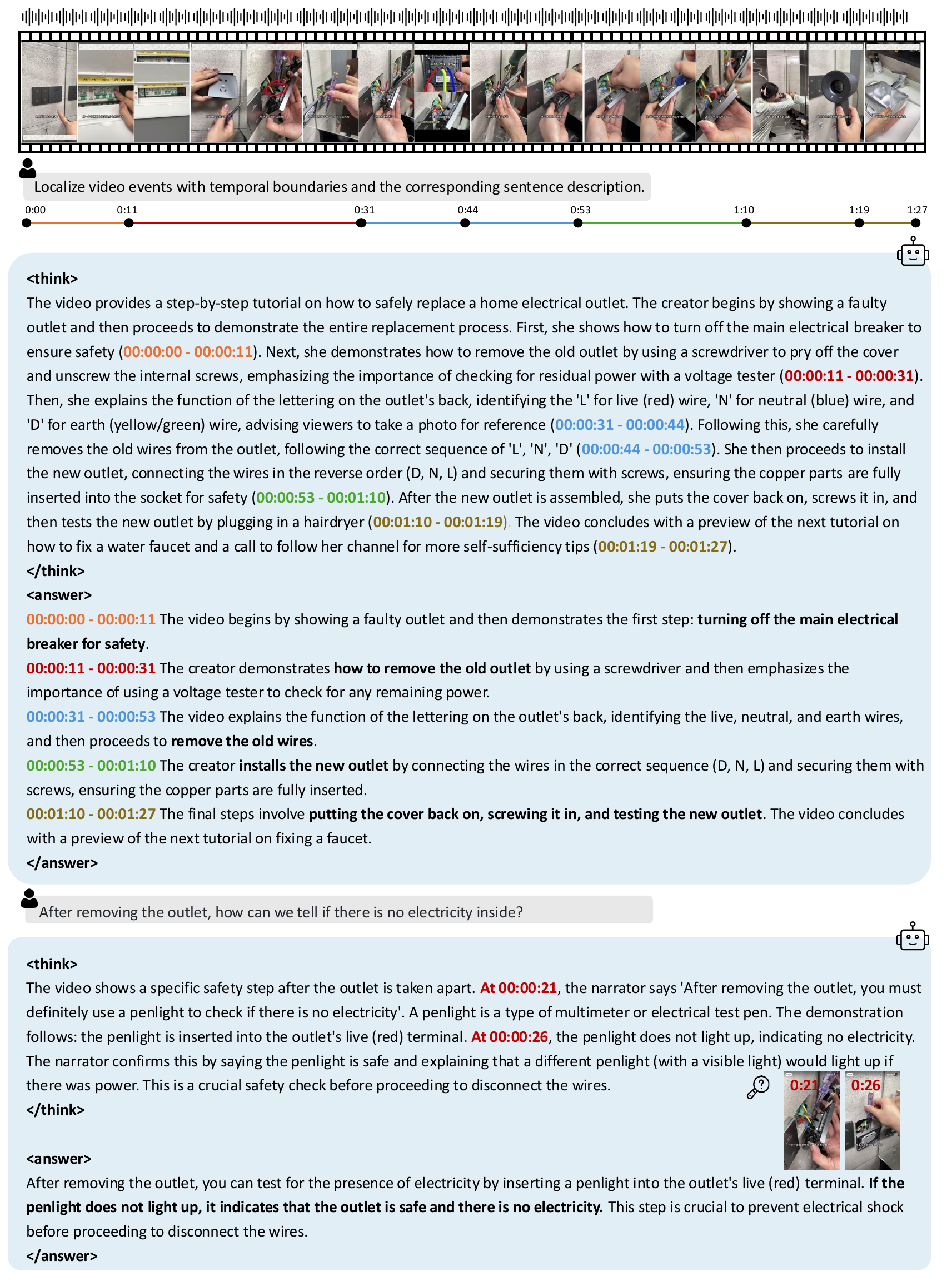}}%
\caption{An example of \modelbase. Given an instructional short video, our model can accurately identify and summarize the content of each step along with the corresponding time spans. For specific questions, the model is also able to locate the relevant time segments within the video, thereby providing precise answers.}
\label{fig:demo1}
\end{figure}

\begin{figure}[H]
\centering
\makebox[\textwidth][c]{\includegraphics[width=1.0\linewidth]{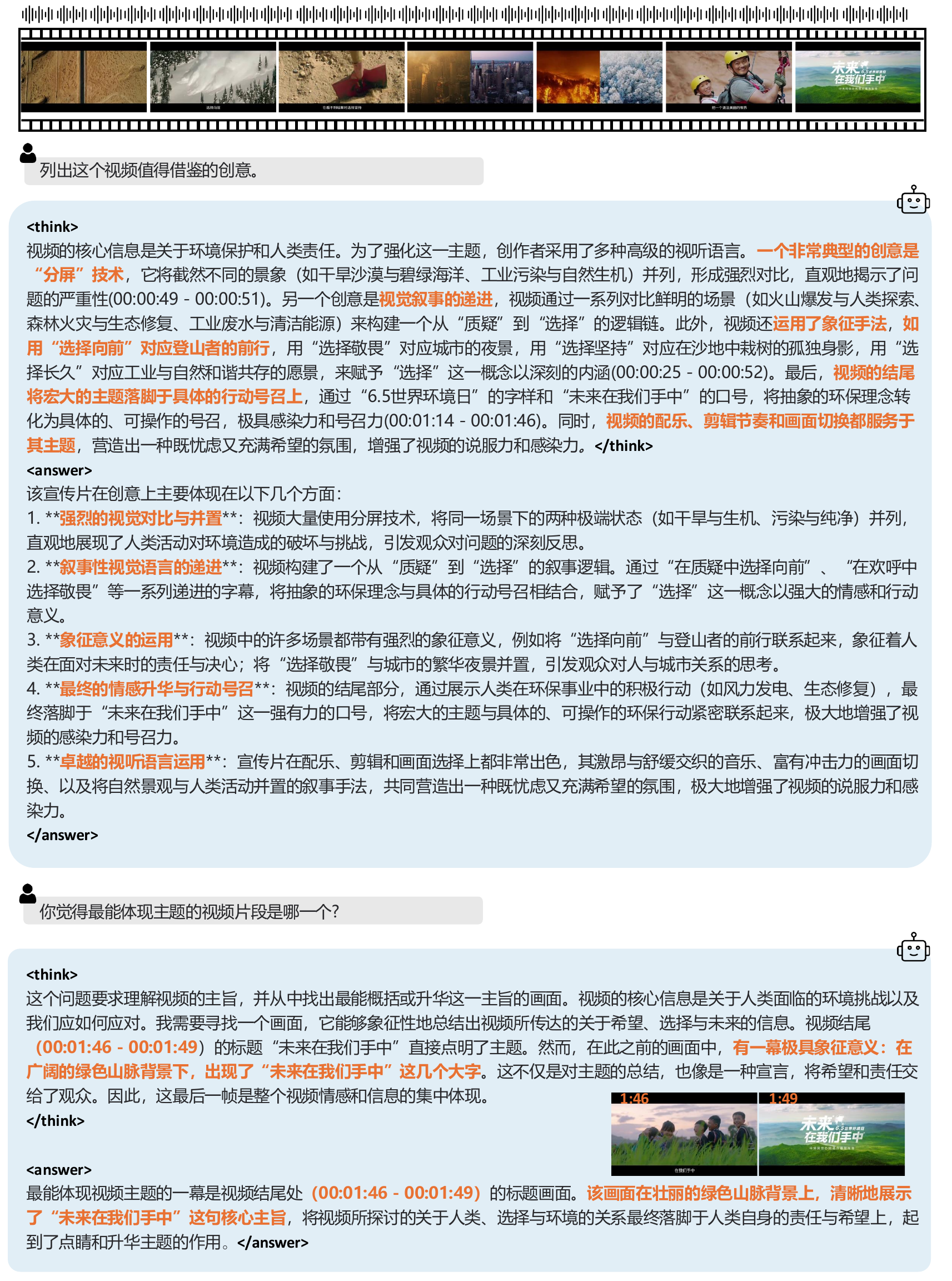}}%
\caption{An example of \modelbase. Given a real-world video with excellent audiovisual quality, our model can analyze the video from visual, auditory, and thematic perspectives, and through reasoning, provide fine-grained segment recommendations.}
\label{fig:demo2}
\end{figure}

\begin{figure}[H]
\centering
\makebox[\textwidth][c]{\includegraphics[width=1.0\linewidth]{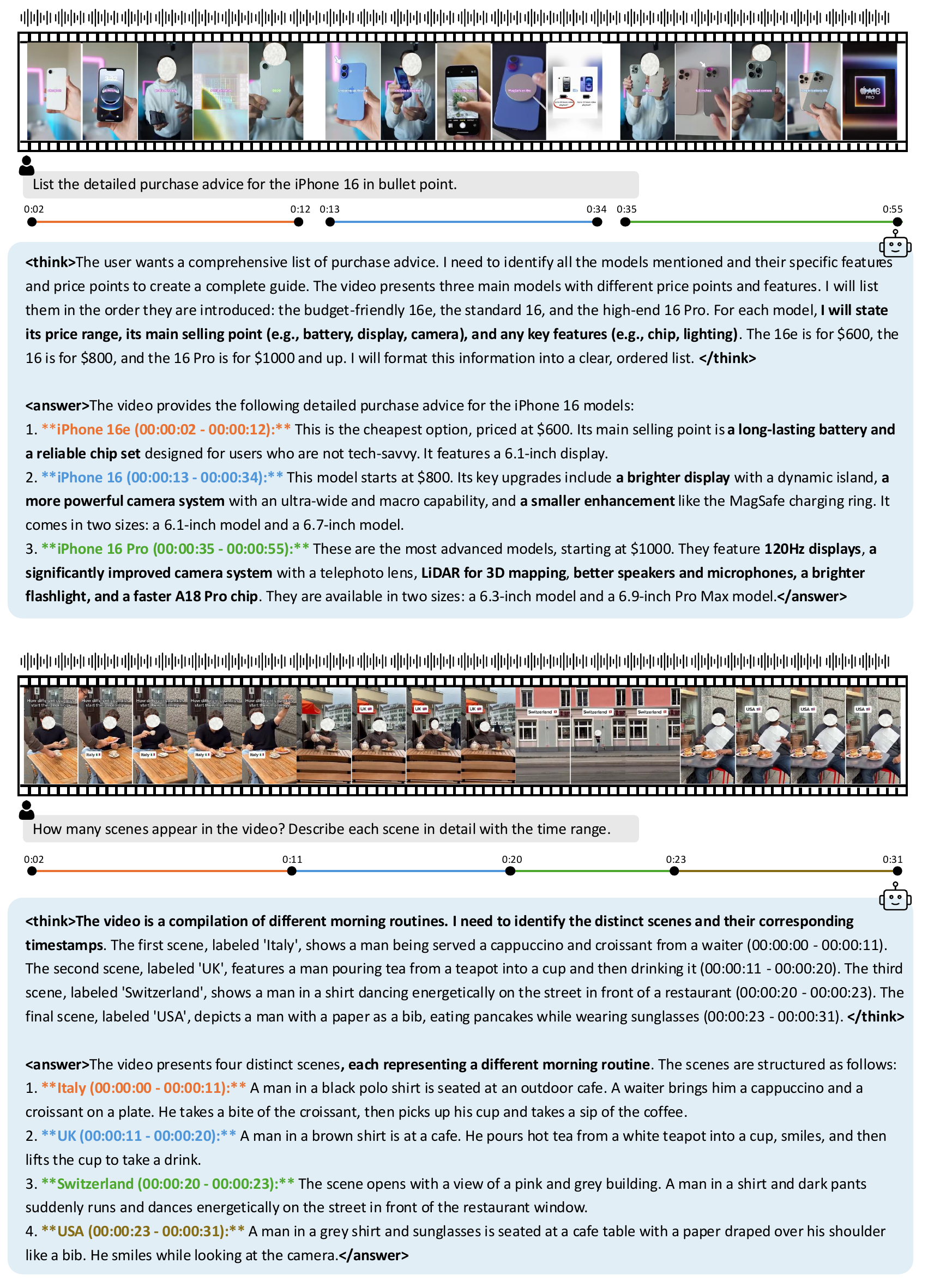}}%
\caption{Examples of \modelbase. Given a review-style short video, the model can extract the characteristics of different products based on both visuals and speech. Given a short video consisting of multiple distinct scenes, our model is able to analyze the transitions between these scenes and accurately discern the main theme.}
\label{fig:demo3}
\end{figure}

To intuitively showcase the advanced capabilities of \modelbase, we present a series of qualitative evaluations on diverse real-world short videos as shown in Fig.~\ref{fig:demo1}, Fig.~\ref{fig:demo2} and Fig.~\ref{fig:demo3}. These examples highlight our model's proficiency in leveraging joint audio-visual reasoning and temporal awareness to achieve deep, structured comprehension, demonstrating its applicability in various real-world scenarios.

\paragraph{\textbf{Joint audio-visual reasoning for complex queries.}} Our model excels at integrating information from multiple modalities to answer complex questions that are unanswerable from a single modality. It is crucial to note that although many short videos feature subtitles, these visual texts can be easily missed or only partially captured at low frame sampling rates. This makes processing the complete audio stream essential for a reliable comprehension of the spoken content, which might otherwise be lost. In the outlet replacement tutorial, when asked how to verify the absence of electricity (Fig.~\ref{fig:demo1}), the model correctly synthesizes the visual action of using a voltage tester with the narrator's spoken instructions to provide a precise and safe procedure. Furthermore, when tasked with summarizing purchase advice from a product review video (Fig.~\ref{fig:demo3}), the model effectively extracts and organizes detailed specifications, prices, and target user profiles for different models, demonstrating its utility in structured information extraction from content where information is distributed across visuals, on-screen text, and narration.

\paragraph{\textbf{Fine-Grained temporal grounding and summarization.}} A core strength of our model is its ability to understand the chronological flow of events. For instance, when analyzing a tutorial video on replacing an electrical outlet (Fig.~\ref{fig:demo1}), \modelbase\ accurately segments the entire process into time-stamped, coherent steps, from turning off the breaker to testing the new installation. This demonstrates its capacity for fine-grained event localization and generating structured, step-by-step summaries. Similarly, for a fast-paced video montage of different morning routines (Fig.~\ref{fig:demo3}), the model successfully identifies and describes each distinct scene with its corresponding time range, showcasing robust scene segmentation. This temporal awareness is crucial for applications like video highlight generation and structured data extraction.

\paragraph{\textbf{High-level thematic and creative understanding.}} Beyond literal descriptions, \modelbase\ demonstrates a remarkable ability for thematic reasoning, which is vital for understanding content focused on emotional expression and viewpoint delivery. When analyzing a real-world promotional video about environmental protection, it can identify sophisticated creative strategies (Fig.~\ref{fig:demo2}), such as the use of ``strong visual contrast'', ``symbolism'' and ``narrative progression'' to convey the message. Moreover, it can pinpoint the single most thematically resonant moment in the video (Fig.~\ref{fig:demo2}), the final shot with the slogan ``The future is in our hands'', showcasing its capacity to grasp the core intent and emotional weight of the content.

In conclusion, these qualitative results validate that \modelbase\ moves beyond surface-level perception to truly understand what happens in a video, when it happens, and why it matters. This deep, structured comprehension makes it a powerful and versatile tool for a wide range of real-world applications.

\subsubsection{Model Comparison} 

\begin{figure}[H]
\centering
\makebox[\textwidth][c]{\includegraphics[width=1.0\linewidth]{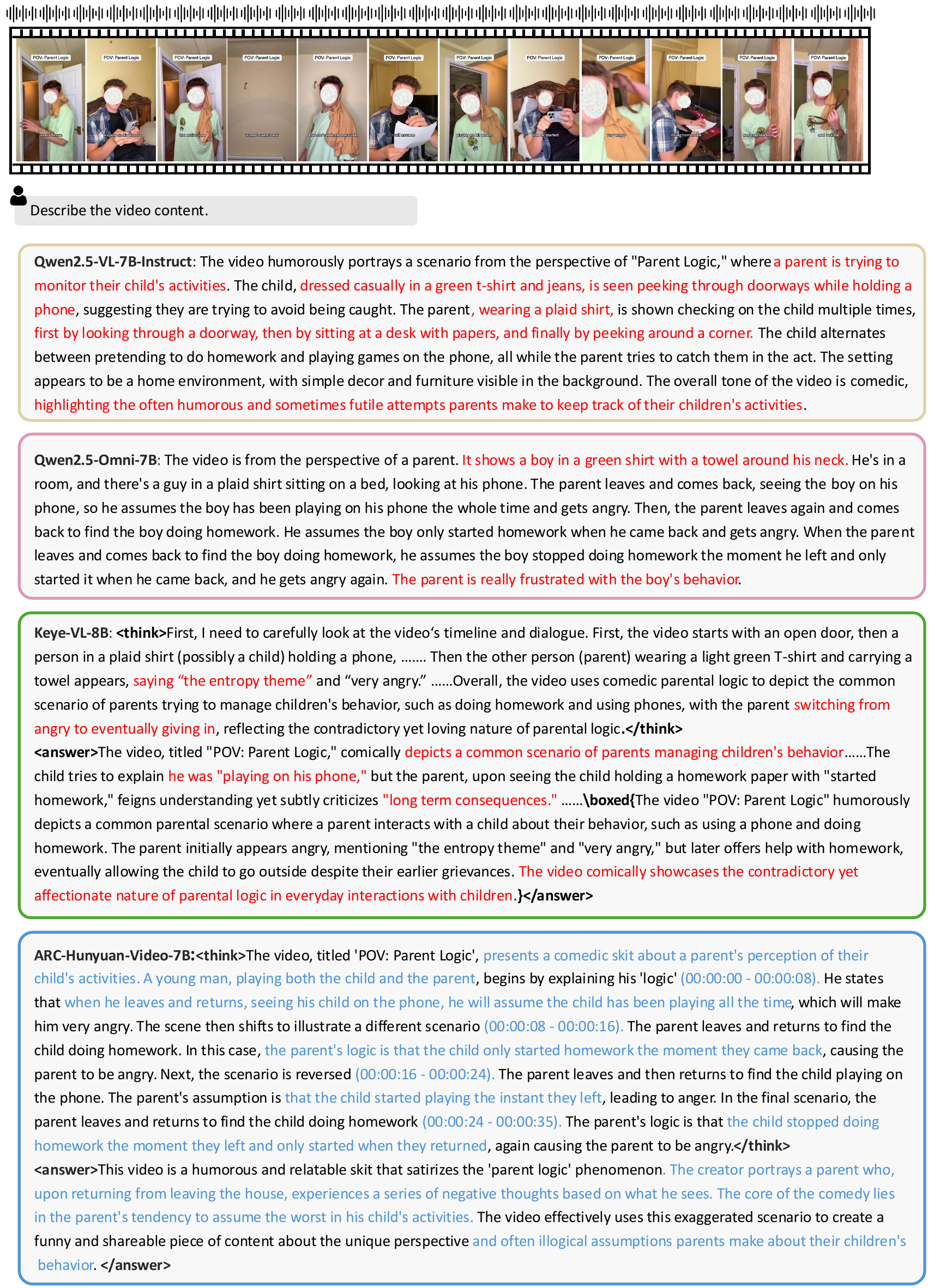}}%
\vspace{-5pt}
\caption{A qualitative comparison between baseline models and our model in understanding short videos where one person plays multiple roles. Our model can accurately identify the events in each scene and provide precise understanding of the main video theme.}
\label{fig:comp1}
\end{figure}

\begin{figure}[H]
\centering
\makebox[\textwidth][c]{\includegraphics[width=1.0\linewidth]{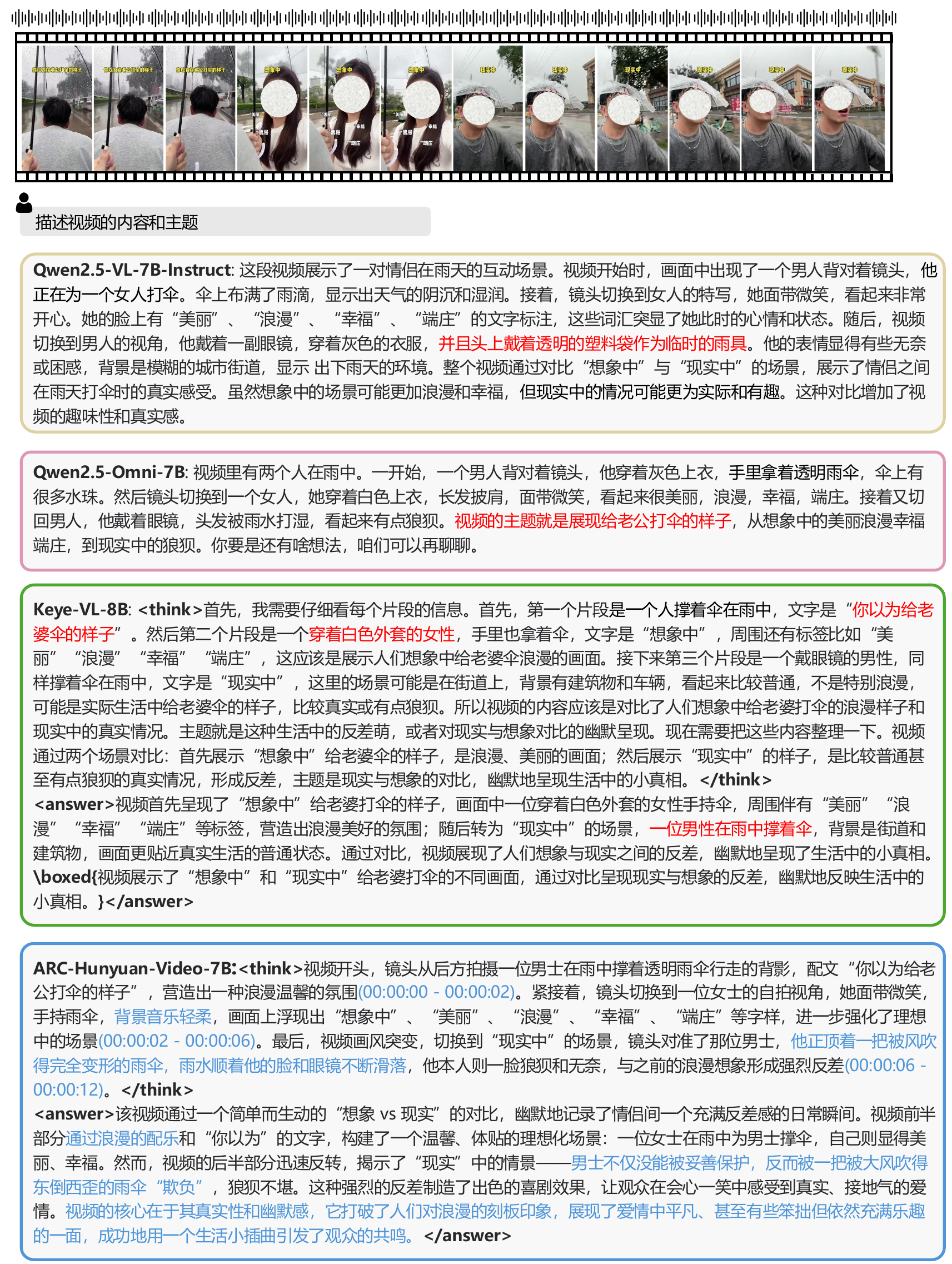}}%
\vspace{-5pt}
\caption{A qualitative comparison between baseline models and our model in understanding short videos with rich visual information. Our model can accurately describe the visual content, analyze the background music, and identify the main theme of the video.}
\label{fig:comp2}
\end{figure}

\begin{figure}[H]
\centering
\makebox[\textwidth][c]{\includegraphics[width=0.9\linewidth]{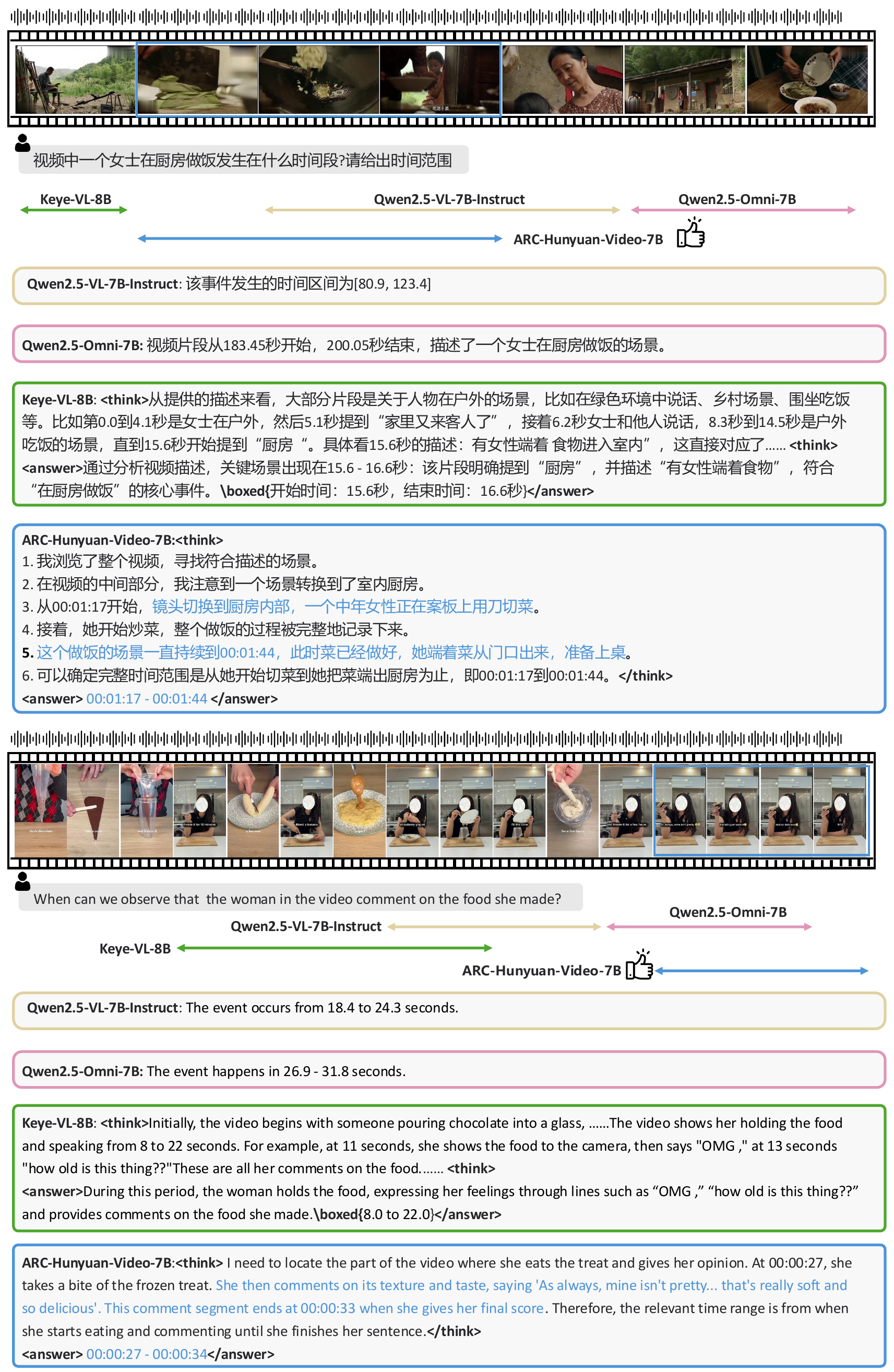}}%
\vspace{-5pt}
\caption{A qualitative comparison between baseline models and our model in the ability of temporal video grounding on real-world videos. Our model can effectively analyze visual and audio cues to accurately determine the start and end times of events.}
\label{fig:comp3}
\end{figure}

To qualitatively assess the capabilities of \modelbase, we conduct a comparative analysis against several baseline models, including Qwen2.5-VL-7B-Instruct~\citep{bai2025qwen2}, Qwen2.5-Omni-7B~\citep{xu2025qwen2}, and Keye-VL-8B-8B~\citep{team2025kwai}. The results, summarized across four representative cases as shown in Fig.~\ref{fig:comp1}, Fig.~\ref{fig:comp2} and Fig.~\ref{fig:comp3}, highlight our model's superior performance in structured video comprehension, particularly in leveraging joint audio-visual reasoning and precise temporal awareness.

\paragraph{\textbf{Superior thematic understanding through audio-visual fusion.}} A primary limitation of video-only models is their inability to process audio, which is often crucial for understanding the context and intent of short videos. In Fig.~\ref{fig:comp1}, a comedic skit titled ``POV: Parent Logic'', the humor and narrative are driven by the audio narration explaining the parent's illogical assumptions. Video-only models like Qwen2.5-VL-7B-Instruct and Keye-VL-8B, deprived of this audio context, misinterpret the visual cues. They describe the physical actions (e.g., a child peeking, a parent checking) but fail to grasp the core comedic premise. Keye-VL-8B even hallucinates non-existent dialogue like ``the entropy theme''. In contrast, \modelbase\ correctly identifies the skit's satirical nature by integrating the audio narration with the visual scenes, accurately summarizing the central theme of ``parent's tendency to assume the worst in his child's activities". While the audio-visual model Qwen2.5-Omni-7B captures the basic events, its summary remains a literal play-by-play, lacking the deeper thematic insight that our model provides.

\paragraph{\textbf{Deeper nuance comprehension in real-world scenarios.}} Beyond just understanding the plot, grasping the nuance and emotional tone is key to short video comprehension. In Fig.~\ref{fig:comp2}, a video contrasting the ``imagination vs. reality" of holding an umbrella for a partner, all models identify the basic visual contrast. However, \model\ excels in capturing the video's intended purpose and emotional impact. Its summary describes the  ``excellent comedic effect'' and how the video ``resonates with the audience" by showing a ``relatable and humorous" side of love. This demonstrates a more profound level of reasoning compared to the baselines, which offer more superficial, descriptive summaries. This ability to understand why a video is engaging is a critical advantage for real-world applications.

\paragraph{\textbf{Enhanced temporal awareness for accurate event grounding.}} The fast-paced nature of short videos makes temporal localization a significant challenge. This weakness is evident in the temporal grounding tasks (Fig.~\ref{fig:comp3}). For the first example, which asks for the time range of a woman cooking, the baseline models produce wildly inaccurate predictions, both completely missing the event. Similarly, in the second example, their predictions for when a woman comments on food are imprecise. \modelbase, however, leverages its explicit temporal awareness, achieved through our timestamp overlay mechanism, and joint visual-audio reasoning, to pinpoint the events with remarkable accuracy. It correctly identifies the cooking scene and the food commentary. This precision demonstrates that our model does not just see what happens, but understands precisely when it happens, a cornerstone of structured video comprehension.

In summary, these qualitative comparisons underscore the effectiveness of \modelbase. By robustly fusing audio-visual information and maintaining a strong sense of temporality, our model overcomes the limitations of video-only and general-purpose multimodal models, delivering a more accurate, nuanced, and structured understanding of real-world short videos.

\subsection{Quantitative Evaluation}
\subsubsection{Evaluation Benchmark} 
\begin{figure}[H]
\centering
\makebox[\textwidth][c]{\includegraphics[width=0.9\linewidth]{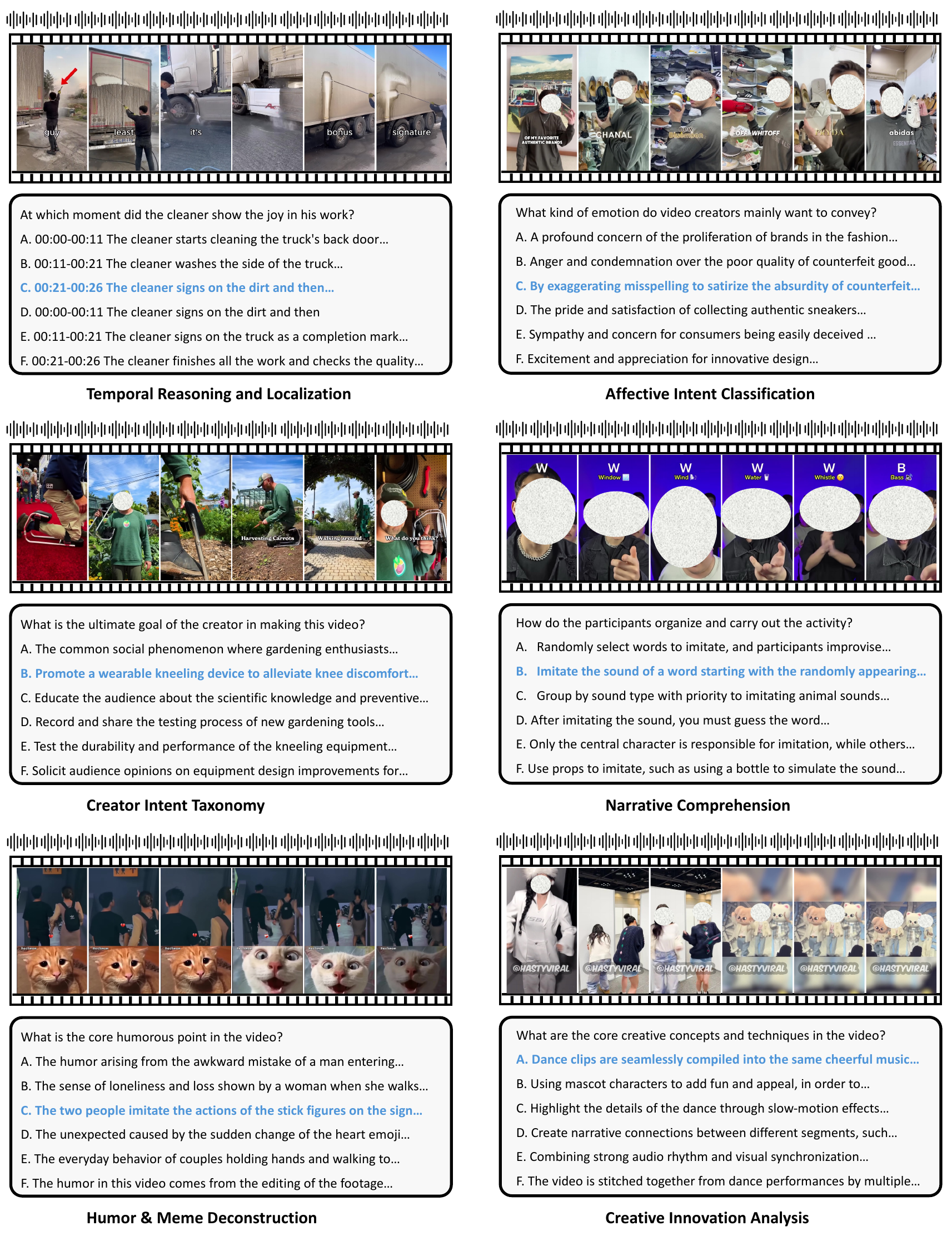}}%
\vspace{-5pt}
\caption{Examples from ShortVid-Bench. The questions, spanning six distinct dimensions, require integrating both visual and audio information for a genuine comprehension of the real-world short videos.}
\label{fig:bench}
\end{figure}

\paragraph{\textbf{Real-world shorts understanding.}}
Existing benchmarks often fall short in capturing the nuanced complexities of user-generated content. To rigorously evaluate our model's ability to understand real-world short videos, we construct a specialized benchmark named \textbf{ShortVid-Bench}. Specifically, we develop an automated pipeline to generate multi-dimensional questions for each video, targeting capabilities that signify a deep, holistic comprehension through integrating both visual and audio cues. These dimensions include: (1) Temporal Reasoning and Localization, (2) Affective Intent Classification, (3) Creator Intent Taxonomy, (4) Narrative Comprehension, (5) Humor \& Meme Deconstruction, (6) Creative Innovation Analysis as shown in Fig.~\ref{fig:bench}. For objective assessment, we employ a multiple-choice question (MCQ) format following previous work~\citep{li2023seed,chen2023egoplan,qiu2024egoplan}. Each question is carefully curated by human annotators who provide the ground-truth answer and design challenging, plausible distractors. Collectively, these dimensions push the evaluation beyond mere descriptive captioning, demanding a genuine comprehension of the video's context, intent, and narrative.

\paragraph{\textbf{Video temporal grounding.}} We further evaluate our model on temporal
video grounding tasks including Charades-STA~\citep{gao2017tall}, which contains 
3,720 long videos capturing indoor human activities for testing, and ActivityNet~\citep{caba2015activitynet}, which comprises 17,031 test samples.

\paragraph{\textbf{General video understanding and reasoning.}} While our primary focus is the structured comprehension of real-world short videos, we also evaluated \modelbase\ on several established general-purpose benchmarks. Specifically, we report performance on (1) MVBench \citep{li2024mvbench}, (2) the multiple-choice task of VCR-Bench \citep{qi2025vcr}, and (3) Video-Holmes \citep{cheng2025video}. The first two benchmarks encompass a mixture of perception and reasoning tasks across various video types, while Video-Holmes is specifically designed to test complex video reasoning, with a focus on suspenseful short films.

\subsubsection{Evaluation Results} 
As shown in Tab.~\ref{tab:video_benchmark}, \modelbase\, achieves the highest accuracy on our proposed ShortVid-Bench, which 
demonstrates its superior ability to comprehend real-world short videos by integrating visual and audio signals with advanced reasoning capabilities.  Furthermore, our model outperforms all baselines in temporal video grounding, which is largely attributed to our strategy of directly overlaying timestamps onto the video frames for enhancing temporal awareness. With limited general-purpose video training data, our model also shows promising results on general video understanding and reasoning benchmarks.

\begin{table*}[!ht]
\centering
\caption{Quantitative evaluation on different benchmarks, which use accuracy as the evaluation metric, except for the grounding tasks, which use mIoU.} 
\label{tab:video_benchmark}
\renewcommand{\arraystretch}{1.5} 
\resizebox{\textwidth}{!}{%
\begin{tabular}{lccccccccc}

\rowcolor{headergray}
\multicolumn{4}{c}{} & 
\multicolumn{1}{c}{\textbf{Real-world Shorts Und}} & 
\multicolumn{2}{c}{\textbf{Temporal Video Grounding}} & 
\multicolumn{3}{c}{\textbf{General Video Und \& Reasoning}} \\

\rowcolor{headergray}
\textbf{Model} & \textbf{fps} & \textbf{\#frames} & \textbf{think} & \textbf{ShortVid-Bench} & \textbf{Charades-STA} & \textbf{ActivityNet} & \textbf{MVBench} & \textbf{VCR-Bench} & \textbf{Video-Holmes} \\
\toprule

\rowcolor{rowblue0}
Qwen2.5-VL-7B-Instruct & 1.0 & 150 &$\times$ &67.8 &46.9 &25.1 &62.9 &\textbf{53.7} &41.6 \\

\rowcolor{rowblue1}
Qwen2.5-Omni-7B & 1.0 & 150 &$\times$ &68.3 &30.5 &13.0 &\textbf{64.8} &51.0 &\textbf{43.9} \\

\rowcolor{rowblue2}
Keye-VL-8B & 1.0 & 150 &$\checkmark$ &53.5 &25.1 &14.9 &35.7 &34.9 &35.7 \\

\rowcolor{rowblue3}
\modelbase & 1.0 & 150 &$\checkmark$ &\textbf{74.3} &\textbf{54.8} &\textbf{41.7} &62.6 &50.5 &40.9 \\
\bottomrule

\end{tabular}%
}
\end{table*}

\subsection{Downstream Application}
To demonstrate the practical utility and adaptability of \modelbase, we conduct supervised fine-tuning on a set of downstream tasks with minimal task-specific data for real-world application scenarios. We map specific supervised fine-tuning tasks to their corresponding real-world applications: (a) Brief Summary for Video Retrieval, (b) Detailed Summary for comprehensive Video Tagging, and (c) Extended Browsing Words for Video Recommendation.

\begin{figure}[H]
\centering
\makebox[\textwidth][c]{\includegraphics[width=1.0\linewidth]{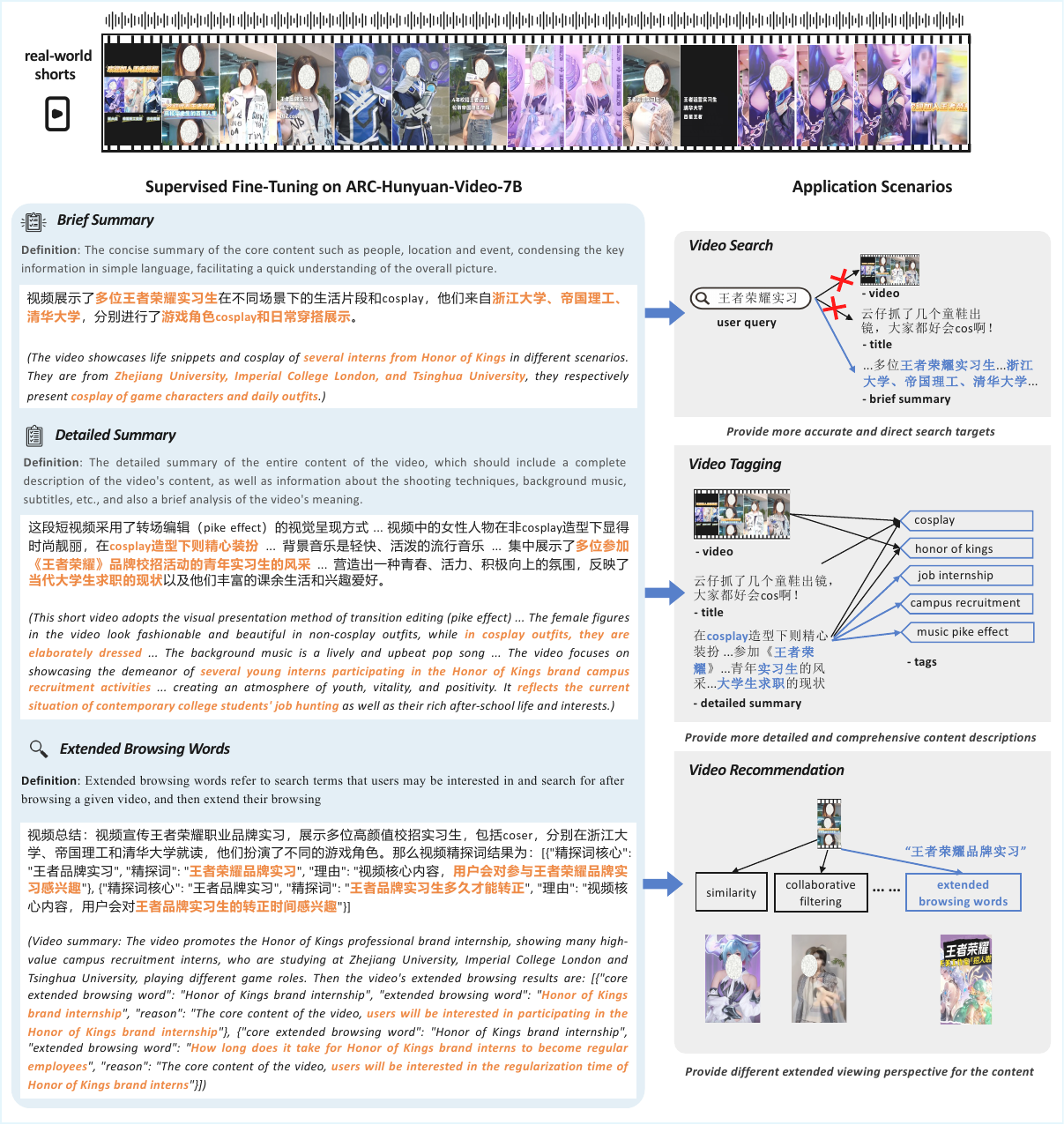}}%
\caption{Demonstration of \modelbase's versatility through minimal fine-tuning for various downstream applications. Specific supervised fine-tuning tasks are mapped to their corresponding real-world scenarios: (a) Brief Summary for Video Retrieval, (b) Detailed Summary for comprehensive Video Tagging, and (c) Extended Browsing Words for Video Recommendation.}
\label{fig:application}
\end{figure}

\subsubsection{Experimental Setup}
Based on common business scenarios, we consider three typical ones as examples. The specific task definitions are as follows:
\begin{itemize}
    \item \textbf{Brief Summary.} The concise summary of the core content such as people, location and event, condensing the key information in simple language, facilitating a quick understanding of the overall picture. Brief summaries can effectively simplify the functional requirements in scenarios such as video retrieval and video aggregation, and transform traditional cross-modal or pure visual analysis into more mature plain text operations.
    \item \textbf{Detailed Summary.} The detailed summary of the entire content of the video, which should include a complete description of the video's content, as well as information about the shooting techniques, background music, subtitles, etc., and also a brief analysis of the video's meaning. Similar to brief summary, detailed summary is also a compromise made to maximize utilizing the efficiency and high quality of plain text retrieval. The difference is that detailed summary pay more attention to the content details of the video and can make more accurate matches to the video content details during retrieval.
    \item \textbf{Extended Browsing Words.} Extended browsing words refer to search terms that users may be interested in and search for after browsing a given video, and then extend their browsing. This is a typical recommendation scenario. Traditional strategies may implement recommendations based on video similarity~\citep{wray2021semantic, fang2021clip2video} or collaborative filtering~\citep{wu2022survey} or association rules~\citep{liao2021investigating, qin2021world} based on user behavior, but extended browsing words given by content-based reasoning can effectively expand the scope of recommendations and have better prospects in terms of cold start and preference prediction.
\end{itemize}

For each of the three tasks, we obtain 1,100 samples by manual annotation, of which 1,000 are used for supervised fine-tuning and the remaining 100 are used for evaluation. To ensure the quality of data annotation, we randomly select 10\% of the labeled data for cross-validation, and the pass rate was greater than 95\%. 

For evaluating the performance of the fine-tuned \model\ from both qualitative and quantitative perspectives, we utilize the following two indicators.

\paragraph{\textbf{Pass Rate (PR).}} We manually define a score scale for evaluating model outputs, which is divided into three grades from low to high, i.e., 0-2 points. Specifically, a score of 0 indicates that the model output contains obvious errors, which are clearly inconsistent with the original video content or task definition; a score of 1 means that the model output has acceptable minor issues that do not affect the understanding of the original video content and do not excessively violate the rules; a score of 2 represents that the model output has no problems at all and fully conforms to the task definition. In actual business operations, we generally believe that a score of 1 or above can be considered a pass.

\paragraph{\textbf{Good vs. Same vs. Bad (GSB).}}~\citep{zou2021pre, zhao2022self, ye2023improving, li2025m2oerank} GSB metric is widely adopted in industry, and it is evaluated by experts judging the superiority or inferiority of results from different sources. Specifically, judges are given two results for a single input: one generated by System A, and the other by its competitor, System B. Crucially, annotators are unaware of which system each result corresponds to. Their task is to determine which result is of higher quality based on an assessment of the overall quality of the output results.

\subsubsection{Implementation Details}
We use the \modelbase\ as the base model and optimize the parameters of the MLP adapter layers and the full LLM backbone with a learning rate of 1e-5 for 3 epochs.
The prompt settings in the training data consist of rule descriptions in natural language. In particular, in the reasoning of extended browsing words, we design a simple thought of chain structure, which outputs the understood video content and then infers possible extended browsing words.

\subsubsection{Experimental Analysis}
\begin{table}[h]
    \centering
    \scriptsize
    \caption{Evaluation results of three tasks, where baseline is the original online model (different for three businesses) and Ours is \modelbase\ after supervised fine-tuning for different tasks. Our model shows marked improvements, with a significantly higher Pass Rate (PR) and a dominant win rate in the GSB (Good vs. Same vs. Bad) human preference comparisons.}
    \label{tab:app_eval}
    \renewcommand{\arraystretch}{1.5}
    \resizebox{\textwidth}{!}{
    \begin{tabular}{lccccccccccccccc}
    \rowcolor{headergray}
     & \multicolumn{5}{c}{\textbf{Brief Summary}} & \multicolumn{5}{c}{\textbf{Detailed Summary}} & \multicolumn{5}{c}{\textbf{Extended Browsing Words}} \\
    \cline{2-16}
    \rowcolor{headergray}
    \textbf{Model} & \textbf{0} & \textbf{1} & \textbf{2} & \textbf{PR} & \textbf{GSB} & \textbf{0} & \textbf{1} & \textbf{2} & \textbf{PR} & \textbf{GSB} & \textbf{0} & \textbf{1} & \textbf{2} & \textbf{PR} & \textbf{GSB}  \\
    \hline
    \rowcolor{rowblue0}
    Baseline (Method A) & 29 & 22 & 49 & 0.71 & & 37 & 44 & 19 & 0.63 & & 18 & 36 & 46 & 0.82 &  \\
    \rowcolor{rowblue1}
    Ours (Method B) & 18 & 4 & 78 & 0.82 & \multirow{-2}{*}{16:4:80} & 26 & 35 & 39 & 0.74 & \multirow{-2}{*}{8:15:77} & 12 & 34 & 54 & 0.88 & \multirow{-2}{*}{14:44:42}  \\
    \bottomrule
    \end{tabular}
    }
\end{table}

\paragraph{\textbf{Main Results.}} As shown in Table~\ref{tab:app_eval}, compared with the current online baseline, the supervised fine-tuned \modelbase\ performs significantly better in the three tasks. Specifically, by comparing the model scores, it can be found that supervised fine-tuned \modelbase\ received fewer 0 points and more 2 points, so the PR is also significantly higher than the baseline, proving that it can better and more correctly understand the video content according to the instructions. At the same time, the relatively large difference in the GSB score based on manual evaluation further proves that even one instance can get a score of 2 points. The supervised fine-tuned \modelbase\ can better meet the preferences of the reviewers and provide a better experience. Going deeper, since both are summary tasks, the main difference between brief summary and detailed summary is the level of detail in the description of the video content. By comparing the PR of the two, it can be found that the detailed summary is more difficult, which is also in line with intuition. After all, the rules of detailed summary are more difficult, and the description of video details is more prone to errors and omissions. Further observation of the distribution of 0, 1, and 2 scores of these two tasks also shows that the number of 0 points in the detailed summary is larger, and the amount of 2 points in the detailed summary is also significantly lower than that of the concise summary, which further indicates that the current model still has room for improvement.

\paragraph{\textbf{Real-world Production Benefits.}} The above three types of supervised fine-tuning tasks are not actually directly output to users for interaction, but can provide indirect support for different user services as intermediate products. For example, the brief summary, as a description of the core content of the video, can directly serve video retrieval services. Therefore, we apply it to the video retrieval of our products as a retrieval target with user query, which significantly improves the user's retrieval experience. Specifically, our retrieval CTR increased by 5.88\%, the landing page consumption time increased by 5.11\%, the video floating layer click CTR increased by 7.26\%, and the long click rate increased by 3.34\%. Similarly, we also applied it to the video aggregation application. After the function was launched, the number of goals per capita increased by 0.63\%, the average QV per capita increased by 0.55\%, and the proportion of satisfied QV increased by 1.77\%.

\begin{figure}[H]
\centering
\makebox[\textwidth][c]{\includegraphics[width=1.0\linewidth]{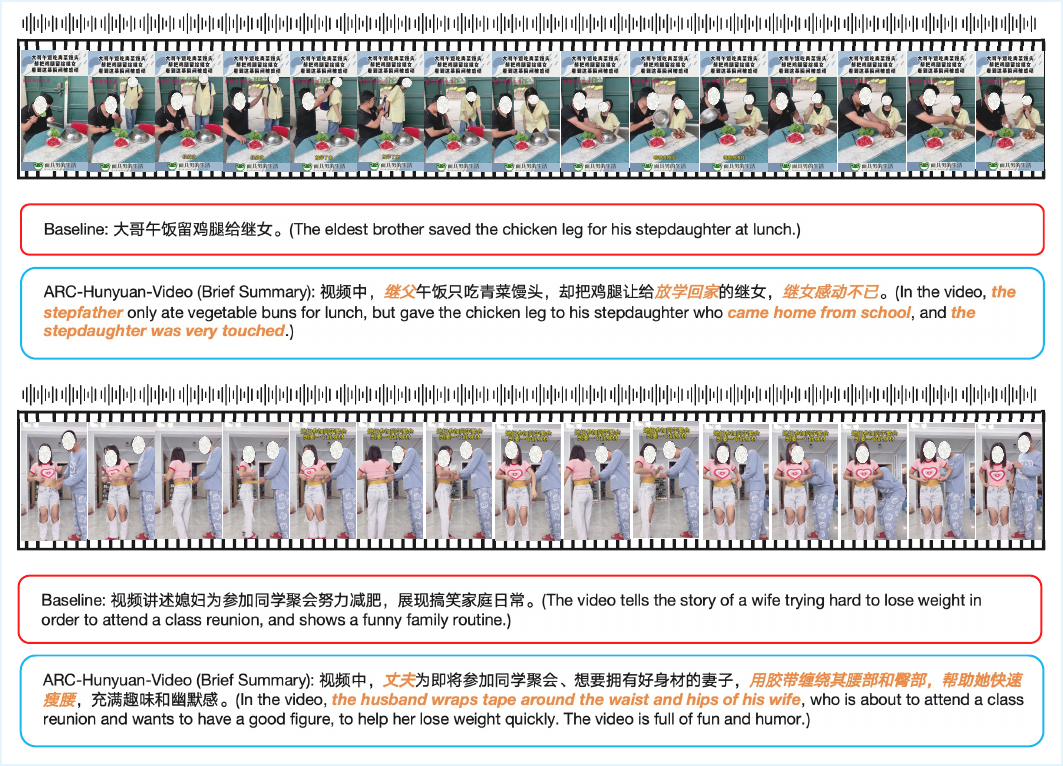}}%
\caption{Examples of brief summary. In the first case, we observe that the fine-tuned \modelbase\ can correctly infer the relationship between the characters in the video, and infer that the person is going home from school through the behavior (carrying a schoolbag into the door). In the second example, the fine-tuned \modelbase\ also correctly identifies the relationship between the characters in the video and describes the complex interaction details between the characters in more detail.}
\label{fig:application_bs}
\end{figure}

\paragraph{\textbf{Case Study.}} To more clearly compare the performance differences after supervised fine-tuning, we provide specific examples for the brief summary in Figure~\ref{fig:application_bs}. By observing these examples, we can find that after supervised fine-tuning, \modelbase\ is able to better incorporate task rules and achieve rule-guided video understanding.

\section{Conclusion}
This paper introduces \model, a powerful multimodal model designed to tackle the challenges of understanding real-world short videos. Faced with the complexity of user-generated content, characterized by dense information, multimodal integration, and rapid pacing, we propose the concept of \textbf{Structured Video Comprehension}, which focuses on fine-grained, temporally-precise understanding of a video's narrative, events, and underlying intent. Built upon on Hunyuan-7B VLM, we adopt an audio encoder for fine-grained audiovisual synchronization and a timestamp overlay mechanism for explicit temporal awareness. This model is trained using a multi-stage strategy on a large-scale dataset of millions of real-world videos, annotated via an automated bootstrapped pipeline. A core finding of our work is that grounding the model in objective tasks with RL is key to unlocking high-quality, subjective understanding. Extensive experiments demonstrate that \model\ achieves state-of-the-art performance on short video comprehension benchmarks and exhibits strong versatility for downstream applications. We believe \model\ represents a significant step towards enabling more sophisticated, in-depth, and practical video-centric AI services, paving the way for a new generation of intelligent video applications.

\bibliographystyle{assets/plainnat}
\bibliography{paper}


\end{document}